\documentclass{article}

\usepackage[preprint]{neurips_data_2024}

\usepackage[utf8]{inputenc} %
\usepackage[T1]{fontenc}    %
\usepackage{hyperref}       %
\usepackage{url}            %
\usepackage{booktabs}       %
\usepackage{amsfonts}       %
\usepackage{nicefrac}       %
\usepackage{microtype}      %
\usepackage{xcolor}         %
\usepackage{amsmath}
\usepackage{amssymb}
\usepackage{mathtools}
\usepackage{amsthm}
\usepackage{tabularx}
\usepackage{multicol}
\usepackage{multirow}
\usepackage{adjustbox}
\usepackage{makecell}
\usepackage{pifont}
\usepackage{subcaption}

\hypersetup{
  colorlinks   = true, %
  urlcolor     = blue, %
  linkcolor    = red, %
  citecolor   = blue %
}
\usepackage{url}

\newcommand{\keypoint}[1]{\noindent\textbf{#1}\quad}

\newcommand{\xmark}{\ding{55}}
\definecolor{darkpastelgreen}{rgb}{0.01, 0.75, 0.24}

\newcommand{\bench}{MIRB}

\title{Benchmarking Multi-Image Understanding in Vision and Language Models: Perception, Knowledge, Reasoning, and Multi-Hop Reasoning}

\author{%
  Bingchen Zhao$^{\star 1}$ \quad Yongshuo Zong$^{\star 1}$ \quad Letian Zhang$^{\star 2}$ \quad Timothy Hospedales$^{1}$\vspace{.3em}
  \\\small $^{\star}$equal technical contribution\vspace{.5em} \\
  \url{https://huggingface.co/datasets/VLLMs/MIRB}\vspace{.3em}\\
  $^1$University of Edinburgh  \qquad $^2$Tongji University
}

\begin{document}

\maketitle
\begin{figure}[h]
    \centering
    \begin{subfigure}[t]{0.49\textwidth}
    \includegraphics[width=\linewidth]{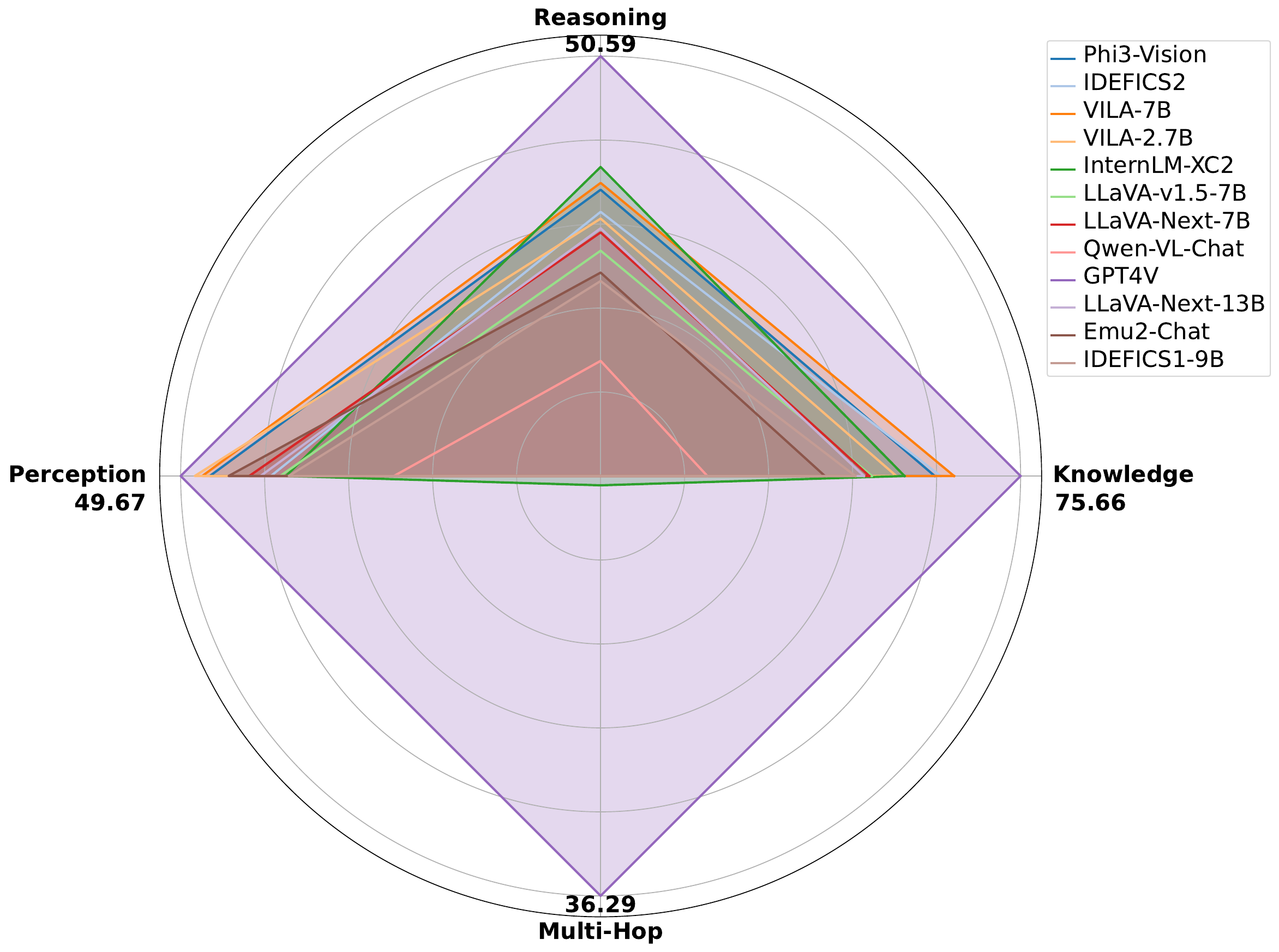}
    \caption{Performance of state-of-the-art large vision-language models on our benchmark.}
    \label{fig:radar}
    \end{subfigure}
    \hfill
    \begin{subfigure}[t]{0.48\textwidth}
    \includegraphics[width=\linewidth]{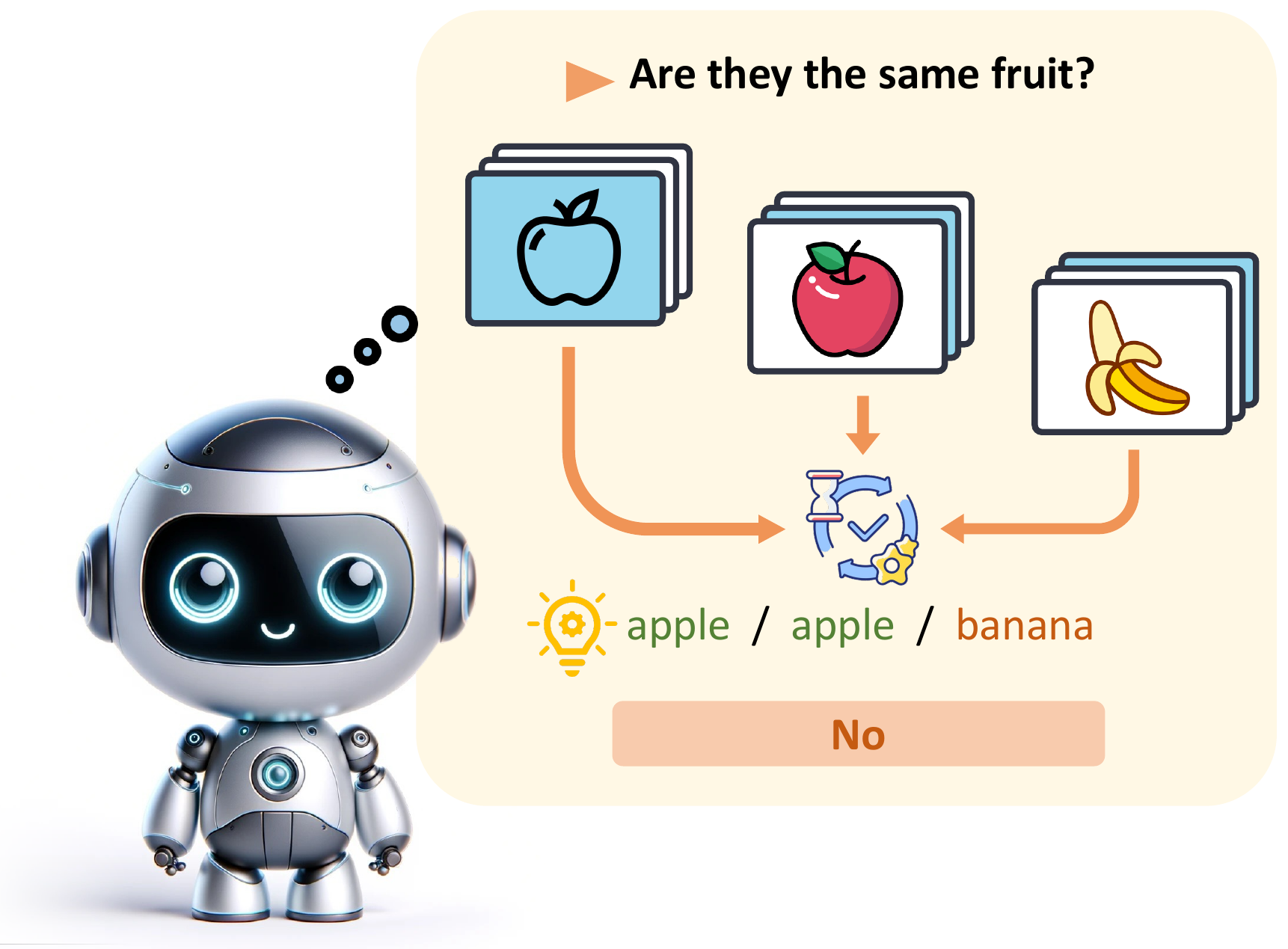}
    \caption{Illustration of one type of perception multi-image reasoning tasks on our benchmark.}
    \label{fig:teaser}
    \end{subfigure}
\end{figure}

\begin{abstract}
The advancement of large language models (LLMs) has significantly broadened the scope of applications in natural language processing, with multi-modal LLMs extending these capabilities to integrate and interpret visual data. 
However, existing benchmarks for visual language models (VLMs) predominantly focus on single-image inputs, neglecting the crucial aspect of multi-image understanding. 
In this paper, we introduce a Multi-Image Relational Benchmark~\textbf{\bench},  designed to evaluate VLMs' ability to compare, analyze, and reason across multiple images. 
Our benchmark encompasses four categories: perception, visual world knowledge, reasoning, and multi-hop reasoning. 
Through a comprehensive evaluation of a wide range of open-source and closed-source models, we demonstrate that while open-source VLMs were shown to approach the performance of GPT-4V in single-image tasks, a significant performance gap remains in multi-image reasoning tasks. 
Our findings also reveal that even the state-of-the-art GPT-4V model struggles with our benchmark, underscoring the need for further research and development in this area. 
We believe our contribution of~\bench~could serve as a testbed for developing the next-generation multi-modal models. 
\end{abstract}

\section{Introduction}

The rise of large language models (LLMs) has enabled numerous groundbreaking applications across various domains. 
For instance, conversational agents like ChatGPT have revolutionized how we interact with technology by providing coherent and contextually relevant responses in natural language~\citep{openai2023gpt4}. 
These models have also shown prowess in tasks such as knowledge-based question-answering~\citep{mmlu}, mathematics~\citep{math}, and code generation~\citep{humaneval}, significantly advancing the state of the art in natural language processing. 
Works have also been done to try to adapt this powerful reasoning ability into real-world workflows by designing agent systems like AutoGPT~\citep{autogpt}.

Large vision language models extend this capability to visual modalities and VLMs are trained to integrate visual inputs to understand and interpret images~\citep{liu2024llavanext, laurenccon2024idefics2, openai2023gpt4, bai2023qwen, liu2023improved}. Accordingly, researchers have been studying the evaluation of the trained VLMs. Currently, most of the evaluation benchmarks primarily focus on understanding and interpreting \textit{single-image} input, either domain-specific (e.g., ScienceQA~\citep{lu2022scienceqa}, MathVista~\citep{lu2024mathvista}, TextOCR~\citep{singh2021textocr}), or aggregated (e.g., MME~\citep{fu2023mme}, SEED-Bench~\citep{li2023seedbench}, MMMU~\citep{yue2023mmmu}, MME~\citep{fu2023mme}).
This narrow focus overlooks the crucial capability of comparing, analyzing, and reasoning across \textit{multiple images}, which is essential for many real-world applications such as comparing different shopping items, analyzing X-ray images from different angles, and understanding a temporal sequences.

In this paper, we address this significant gap by designing a benchmark specifically aimed at multi-image evaluation. 
Our benchmark comprises four distinct categories of multi-image understanding: perception, visual world knowledge, reasoning, and multi-hop reasoning. 
Each category includes a range of tasks that necessitate comparison of multiple input images to derive solutions. 
These tasks are designed to push the boundaries of current VLM capabilities and to provide a more comprehensive evaluation of their reasoning abilities.

Our comprehensive evaluation of open-source and closed-source models on this benchmark reveals significant insights. 
While open-source visual language models like LLaVA perform comparably to GPT-4V in single-image reasoning and question answering, a substantial performance gap persists in multi-image reasoning tasks. 
Furthermore, even the state-of-the-art closed-source GPT-4V struggles to achieve high performance on our benchmark, highlighting the complexity and challenges inherent in multi-image reasoning.
We believe that our benchmark could contribute to the development of open-source multimodal models that can comprehend and reason over multiple images at once to enable richer application scenarios.

To summarize, the contributions of this paper are three-fold: Firstly, we introduce a comprehensive benchmark~\bench~for evaluating diverse facets of multi-image understanding, filling a critical gap in the evaluation of visual language models. Second, we provide a detailed evaluation of both open-source and closed-source models, highlighting the current limitations and performance discrepancies in multi-image reasoning. Third, our findings underscore the challenges and potential areas for improvement in the development of visual language models capable of handling and reasoning over multiple images, offering a roadmap for future research and development in this domain.

\section{Related Work}
\keypoint{Large Vision-Language Models.}
With the development of large language models (LLMs)~\citep{openai2023gpt4, touvron2023llama, reid2024gemini}, researchers have been advancing the capabilities of large vision-language models (LVLMs), which are built on LLMs with an additional visual encoder and connection module. Through visual instruction fine-tuning, researchers have trained LVLMs to understand multimodal image-text inputs, demonstrating strong capabilities in perception, reasoning, and more~\citep{liu2023visual, liu2024llavanext, bai2023qwen, abdin2024phi3, dong2024internlmx2,zhao2023tuning,zhao23mug}.

However, most LVLM training data consists of single image-text pairs or pure-text data~\citep{shang2024llava, dai2023instructblip, chen2023minigpt}, which limits their ability to understand inputs containing multiple images. Recently, researchers have focused on training LVLMs to comprehend multiple images using interleaved image-text corpora such as MMC4~\citep{zhu2023multimodal} or OBELICS~\citep{laurenccon2024obelics, lin2024vila, laurenccon2024idefics2, dong2024internlmx2}. However, evaluations of these models' ability to process multiple images have predominantly remained qualitative.

\keypoint{Evaluation of Large Vision-Language Models.}
Evaluation of vision language models has been an important topic for a long period. Traditional benchmarks often focus on one single task such as optical character recognition (OCR)~\citep{ocrvqa,singh2021textocr}, image captioning~\citep{lin2014microsoft}, or visual question answering~\citep{VQA, lu2022scienceqa, gurari2018vizwiz,zong2023fool,tu2023how,Zhang_2024_CVPR}.
With the popularity of LVLMs, researchers have been developing aggregated benchmarks to evaluate different aspects of the developed models, such as MME~\citep{fu2023mme}, MM-Vet~\citep{yu2023mmvet}, MMBench~\citep{liu2023mmbench}, SeedBench~\citep{li2023seedbench}, and MMMU~\citep{yue2023mmmu}. However, these benchmarks are designed to only test the ability of understand \textit{single} image inputs. 

Although less explored, there are also efforts in multi-image evaluation. For example, Q-Bench~\citep{wu2023q} is proposed to evaluate the low-level perception of LVLMs by comparing multiple image inputs. Memontos~\citep{wang2024mementos} evaluates the temporal understanding of the image sequences. Seed-Bench-2~\citep{li2023Seed-bench-2} mainly focuses on existing VQA-based evaluations and lacks the evaluation of reasoning over multiple images.
The DEMON benchmark~\citep{li2023fine} also present several subset of questions requiring multi-image reasoning, where the main focus is on evaluating the demonstrative instruction following abilities of LVLMs.
Most related, and concurrent, to our work is BLINK~\citep{fu2024blink}. As shown in Table~\ref{tab:benchmark_compare}, We cover a more diverse range of tasks and evaluation dimensions, offering a comprehensive benchmark for the evaluation of modern LVLMs.

\section{Dataset and Task Categories}
We consider four evaluation dimensions in our benchmark, including multi-image reasoning, visual world knowledge, perception, and multi-hop reasoning. 
Within each of these dimensions, we design questions that are only solvable by cross-comparing multiple image inputs.
The detailed design of each dimension is explained below.

\begin{figure}[t]
    \centering
    \includegraphics[width=.92\linewidth]{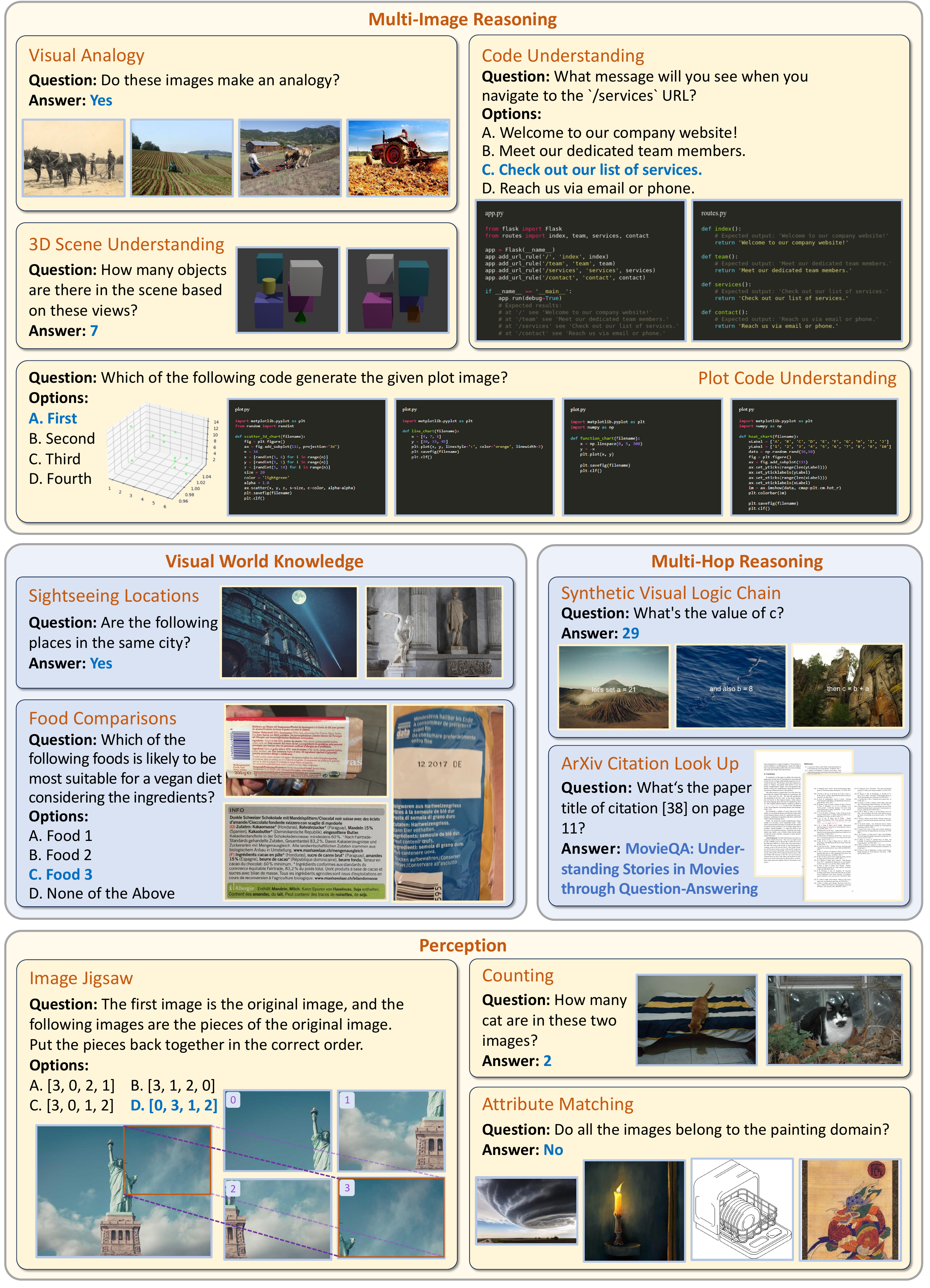}
    \caption{Illustrative examples of~\bench tasks.}
    \label{fig:interaction}
\end{figure}

\subsection{Design of Subtasks}
In this section, we briefly introduce the design of each subtask in \bench. 
The source of images or the generation process of the images are detailed in the supplementary.

\keypoint{Multi-Image Reasoning.}~~~
In this dimension, we evaluate how well models comprehend multiple images, and reason across them to reach the final answer. We include the following types of questions:
\begin{itemize}
    \item Code understanding. In a real-world scenario where a programmer needs to figure out the output of a program, they often need to cross-compare codes from different files to understand the execution trace of the code. In this setting, we collect a set of programs for real-world programming tasks by leveraging the example codes from popular Python libraries for scientific computing, web frameworks, etc\footnote{Complete list of code example sources will be in the supplementary.}. These codes are captured as screenshots and the model is evaluated on how well it understands the code.
    \item Plot code understanding. A second scenario for multi-modal code understanding involves comparing code to the graphical objects that the code generates. Specifically, we generate a set of code that produces plots in various formats including barplot, piechart, etc. We take code examples from the Matplotlib library to build these examples. The model is then given the plotted figure and screenshots of source code, and is tasked to report which source code listing generates the given plot.
    \item Visual analogy. We also consider making visual analogies as a task that requires the model to compare different images. Visual analogies follow the form of what is B' to B given A' to A. The model will need to understand the transformation between A' and A and then apply this to B to be able to reach the correct answer. We take the dataset curated by~\citet{bitton2023vasr} to create this subtask.
    \item 3D scene understanding. Understanding 3D scenes and objects given multiple 2D images is a key task and challenge for robots and embodied agents. As prototypical examples of such tasks, we generate views for synthetic 3D scenes where certain information is only accessible by comparing the different 2D views. For example, counting the total number of objects in the scene given that each individual view has occlusions. We generate the data for this subset by using the Blender engine\footnote{https://docs.blender.org/}.
\end{itemize}

\keypoint{Visual World Knowledge.}~~~
In real-world usage of LVLMs, many popular use cases require analysing visual inputs in the context of world knowledge, for example, determining which food product is more suitable for people with diabetes from the ingredients list. These tasks require combining perception with prior world knowledge. We collect two types of questions within this dimension to understand whether LVLMs can exploit world knowledge in the context of multiple image inputs. 
\begin{itemize}
    \item Sightseeing. We collect data on sightseeing locations in major cities around the world, and generate questions to ask whether or the locations from multiple images is within the same city or not, and also questions asking which city is the presented locations in. For images in this subset, we query the Pixabay API\footnote{https://pixabay.com/api/docs/} and then manually filtered the resulting images.
    \item Food comparisons. The second task is to compare the ingredient list of multiple food products against the criterion given in the text prompt. We collect the image of the food ingredient list from OpenFoodFact~\citep{openfoodfact}, and generate questions that require comparing the food ingredient images.
\end{itemize}

\keypoint{Perception.}~~~
This dimension evaluates how the model perceives the multiple visual inputs. 
We include tasks to measure the ability to perceive and recognize visual input across several images.
\begin{itemize}
    \item Image jigsaw. In this task, the model is provided with an original image and several image patches generated from the original image. The task is to select the correct permutation that put the image patches back to the original image. This requires the model to reason about the multiple image patches and their comparative spatial locations and appearances. For this task, we use random images queried from picsum \footnote{https://picsum.photos/}.
    \item Counting. Another task is to count the number of a certain category of object across multiple images. To generate the correct number of objects, the model needs to perceive and recognize the object across all the image inputs. We use the bounding box annotations from the MS-COCO~\citep{lin2014microsoft} dataset to create this subset.
    \item Attribute matching. As another perception task, we ask the model to match object attributes between multiple images by using questions like ``are the objects in all the input images rendered in the same artistic style?". The model is required to recognize these attributes within each of the images and then associate them. To generate questions that cover comparisons of the different attributes of the image, we use the ImageNet-R~\citep{hendrycks2020many} dataset which contains annotations on the object categories and the artistic style of the images.%
\end{itemize}

\keypoint{Multi-Hop Reasoning.}~~~
Another dimension for evaluating reasoning with multiple images is to learn to associate the content within each of the images to perform the final reasoning. 
For example, information can be provided on one image input, and then transformed in another, forcing the model to attend to different images in the inputs to perform reasoning.
\begin{itemize}
    \item Synthetic visual logic chain. We designed a procedure for generating a chain of images which requires the model to perform reasoning based on the content of each image to reach the final answer. We take random images from picsum and put the text like ``let variable a equal to 1" and ``Set variable b to a+1" to different images, and then ask the model about the value of the variable b. With this design, the model cannot generate a correct answer if it is only able to understand information within one image. 
    \item ArXiv paper citation look up. One real-world example of this relational reasoning can be the process of reading arXiv papers and find the link to references from the content pages. We collect a set of papers and generate questions like ``what is the title of citation [69] on page 15" given the screenshot of the paper. Similar to the synthetic setting, this setting requires the model to form a relation between images to perform reasoning.
\end{itemize}

\begin{table}[t]
  \caption{Comparison with previous multi-image benchmarks. \bench~not only covers all four dimensions for evaluating multi-image reasoning, but it is also built with images sourced independently rather than multiple frames in the same video.}
  \label{tab:benchmark_compare}
  \centering
  \resizebox{0.8\textwidth}{!}{ 
  \setlength\tabcolsep{6pt}
  \begin{tabular}{lcccc}
    \toprule
    \textbf{Benchmark} &  \textbf{Reasoning} & \textbf{Visual Knowledge} & \textbf{Perception} & \textbf{Multi-Hop} \\
    \midrule
    NLVR2 & \textcolor{darkpastelgreen}{\checkmark} & \textcolor{red}{\xmark} & \textcolor{red}{\xmark} & \textcolor{red}{\xmark} \\
    Q-Bench &  \textcolor{red}{\xmark} & \textcolor{red}{\xmark} & \textcolor{darkpastelgreen}{\checkmark}  & \textcolor{red}{\xmark} \\
    Memontos & \textcolor{red}{\xmark} & \textcolor{red}{\xmark} & \textcolor{red}{\xmark} & \textcolor{darkpastelgreen}{\checkmark}   \\
    SEED-Bench-2 & \textcolor{red}{\xmark} & \textcolor{darkpastelgreen}{\checkmark}& \textcolor{red}{\xmark} & \textcolor{darkpastelgreen}{\checkmark}\\
    DEMON & \textcolor{darkpastelgreen}{\checkmark}  &  \textcolor{darkpastelgreen}{\checkmark} & \textcolor{red}{\xmark} & \textcolor{red}{\xmark}  \\
    BLINK & \textcolor{darkpastelgreen}{\checkmark}  & \textcolor{red}{\xmark} &\textcolor{darkpastelgreen}{\checkmark}  &\textcolor{red}{\xmark} \\
    \midrule
    Our \bench{} & \textcolor{darkpastelgreen}{\checkmark}  &\textcolor{darkpastelgreen}{\checkmark}  &\textcolor{darkpastelgreen}{\checkmark}  &\textcolor{darkpastelgreen}{\checkmark}  \\
  \bottomrule
  \end{tabular}
  }
\end{table}

Our benchmark design focuses on testing models' ability to reason with multiple image inputs.
In Table~\ref{tab:benchmark_compare}, we compare \bench~with other benchmarks that can be used for evaluating reasoning on multiple images, the novelty of \bench~not only lies in that \bench~covers four different dimensions of multi-image reasoning, but also is that the images within \bench~are sourced independently rather than directly using the frames from a video which could contain many redundancy within all the input images.
Table~\ref{tab:benchmark_statistics} demonstrates the statistics of our benchmark. For each of the questions in our benchmark, at least 2 images need to be processed to perform the reasoning task. In the case of multi-hop reasoning, the number of images needed for reasoning can go up to 42. 
On average, to answer one question in our benchmark requires the model to process 3 images.
Table~\ref{tab:benchmark_statistics} also presents the question types and the evaluation metrics of each evaluation dimension in our benchmark.

\begin{table}[h]
  \caption{Detailed statistics of the proposed benchmark.}
  \label{tab:benchmark_statistics}
  \centering
  \resizebox{1.0\textwidth}{!}{ 
  \setlength\tabcolsep{6pt}
  \begin{tabular}{lcccccc}
    \toprule
    \textbf{Subsets} &   \textbf{\# Samples} & \textbf{\# Images Range} & \textbf{Avg. Image } &\textbf{Question Type} &\textbf{Metrics} \\
    \midrule
    Reasoning & 254 & [2, 7] & 4.64  & MCQ & Accuracy\\
    Knowledge & 202 & [2, 3] & 2.26  & MCQ & Accuracy\\
    Perception & 350 & [2, 5] & 3.42 & MCQ \& Free-form & Accuracy\\
    Multi-Hop & 119 & [2, 42] & 5.66  & MCQ \& Free-form & Accuracy\\
    \midrule
    Total &  925  & [2, 42]  & 3.78  &  MCQ \& Free-form  &    Accuracy\\
  \bottomrule
  \end{tabular}
  }
\end{table}

\section{Experiments}
In this section, we present the evaluation results of different models on our proposed benchmark.

\keypoint{Models} 
We comprehensively evaluate 12 state-of-the-art VLLMs of various model families and sizes (3-37B) including  LLaVA-v1.5 (7B)~\citep{liu2023improved}, LLaVA-Next (7B/13B)~\citep{liu2024llavanext}, Qwen-VL-Chat (9B)~\citep{bai2023qwen}, InternLM-XComposer2 (7B)~\citep{dong2024internlmx2}, VILA (2.7B/7B)~\citep{lin2024vila}, Emu2-Chat (37B)~\citep{sun2023emu2}, IDEFICS1 (9B)~\citep{laurenccon2024obelics}, IDEFICS2 (8B)~\citep{laurenccon2024idefics2}, Phi-3-Vision (4B)~\citep{abdin2024phi3}, and GPT4-V~\citep{openai2023gpt4}. 
The training data of LLaVA family models contains only single-image pairs, while the others contain interleaved image-text data. 
To ensure reproducibility, we use greedy decoding for the open-sourced models and use the ``2024-05-13'' API version for GPT4-V.

\keypoint{Performance}
In Table~\ref{tab:results}, we present the overall performance comparison of all the models we have tested.
We establish the random chance performance baseline by assuming the model generates uniform random choices when answering multiple-choice questions. And when the question type is free-form, the random model gets an accuracy of zero.
We can see that on the dimension of perception, the best open-source model can perform close to the performance of the state-of-the-art close-sourced GPT4-V model.
However, on other dimensions like reasoning, these open-source models all lag behind GPT4-V by a great margin. 
Most interestingly, on the dimension of visual world knowledge and multi-hop reasoning, \textit{no open-source model can reliably outperform the random chance baseline}, indicating a vast space for exploring the design of VLMs that are able to perform these tasks.

\begin{table}[h]
  \caption{Performance comparison of all models on the four dimensions of \bench.}
  \label{tab:results}
  \centering
  \resizebox{1.0\textwidth}{!}{ 
  \setlength\tabcolsep{6pt}
  \begin{tabular}{lllllllll}
    \toprule
    \textbf{Models} & \textbf{Reasoning}& \textbf{Knowledge} & \textbf{Perception} & \textbf{Multi-Hop} & \textbf{Average} \\
    \midrule
    Random Chance & 20.80 & 37.62 & 21.42 & 0.00 & 23.02  \\
    \midrule
    LLaVA-v1.5-7B & 48.86 & 27.14 & 37.89 & 0.00 & 28.47 \\
    LLaVA-Next-7B & 48.40 & 29.35 & 41.56 & 0.00 & 29.83 \\
    LLaVA-Next-13B & 48.44 & 29.85 & 40.22 & 0.00 & 29.38 \\
    \midrule
    Qwen-VL-Chat & 19.23 & 13.87 & 24.44 & 0.00 & 14.38 \\
    InternLM-XComposer2 & 54.74 & \underline{37.23} & 37.22 & 0.81 & 32.50 \\
    VILA-2.7B  & 53.27 & 31.01 & \underline{48.33} & 0.00 & 33.15 \\
    VILA-7B & \underline{63.66} & 35.31 & 47.11 & 0.00 & \underline{36.52} \\
    Emu2-Chat & 40.40 & 24.51 & 44.00 & 0.00 & 27.23 \\
    IDEFICS1-9B & 45.89 & 23.49 & 36.89 & 0.00 & 26.57 \\
    IDEFICS2-8B & 61.26 & 31.83 & 39.00 & 0.00 & 33.02 \\
    Phi-3-Vision & 60.19 & 34.49 & 46.22 & 0.00 & 35.23 \\
    \midrule
    GPT-4V & \textbf{75.66} & \textbf{50.59} & \textbf{49.67} & \textbf{36.29} & \textbf{53.05} \\
  \bottomrule
  \end{tabular}
  }
\end{table}

\section{Analysis}
In this section, we conduct further analysis to better understand multi-image evaluation. Specifically, we ask the following questions:
\begin{itemize}
    \item \textbf{Q1:} Is our benchmark solvable by using only one image?
    \item \textbf{Q2:} Is concatenating multiple images into one image easier for LVLMs to understand?
    \item \textbf{Q3:} Can LVLMs understand text inputs better than the image inputs?
    \item \textbf{Q4:} Will Chain-of-Thought prompting help the reasoning over multiple images?
\end{itemize}

\subsection{Our benchmark is Not Solvable with Only One Image} Prior multi-modal benchmarks have suffered from the issue of surprising shortcuts providing trivial solutions \citep{goyal2017vqaMatter}. This question therefore validates the goal of our benchmark by testing the null hypothesis that a single image solution to our task exists. To answer this question, we conduct the experiment on 3D scene understanding, code understanding, and synthetic visual logic chain. 
Other subsets are naturally not solvable with one image as they require comparison between images or the options are different images. 
We average the performance on these three subsets and present the comparison of using all image inputs and using only one image input in Figure~\ref{fig:single}. The results show that models require utilising multiple images to achieve reasonable performance.

\subsection{Concatenated Images} 
\begin{figure}
    \begin{subfigure}[t]{0.49\textwidth}
    \includegraphics[width=\linewidth]{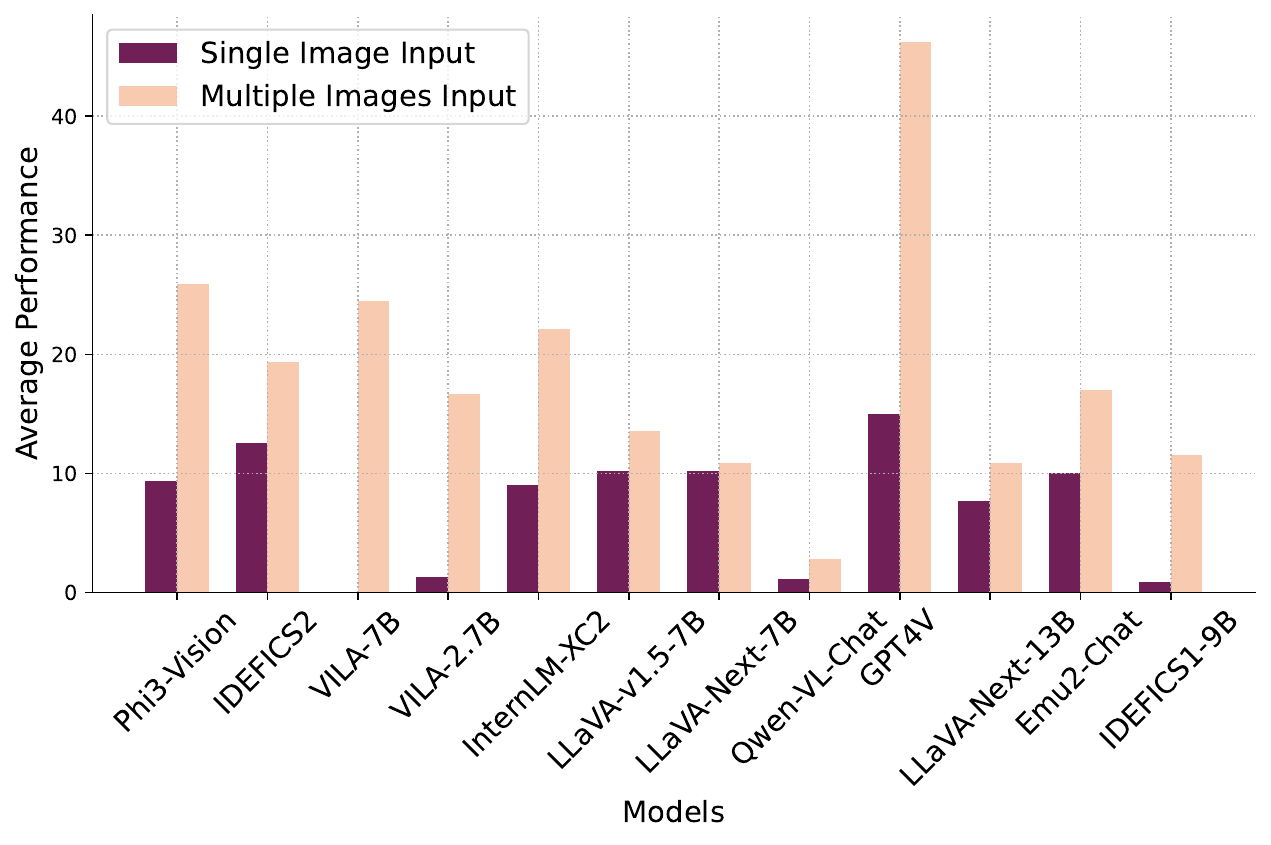}
    \caption{Comparisons of single- and multi-image inputs. Multiple inputs are necessary for good performance.}
    \label{fig:single}
    \end{subfigure}
    \hfill
    \begin{subfigure}[t]{0.49\textwidth}
    \includegraphics[width=\linewidth]{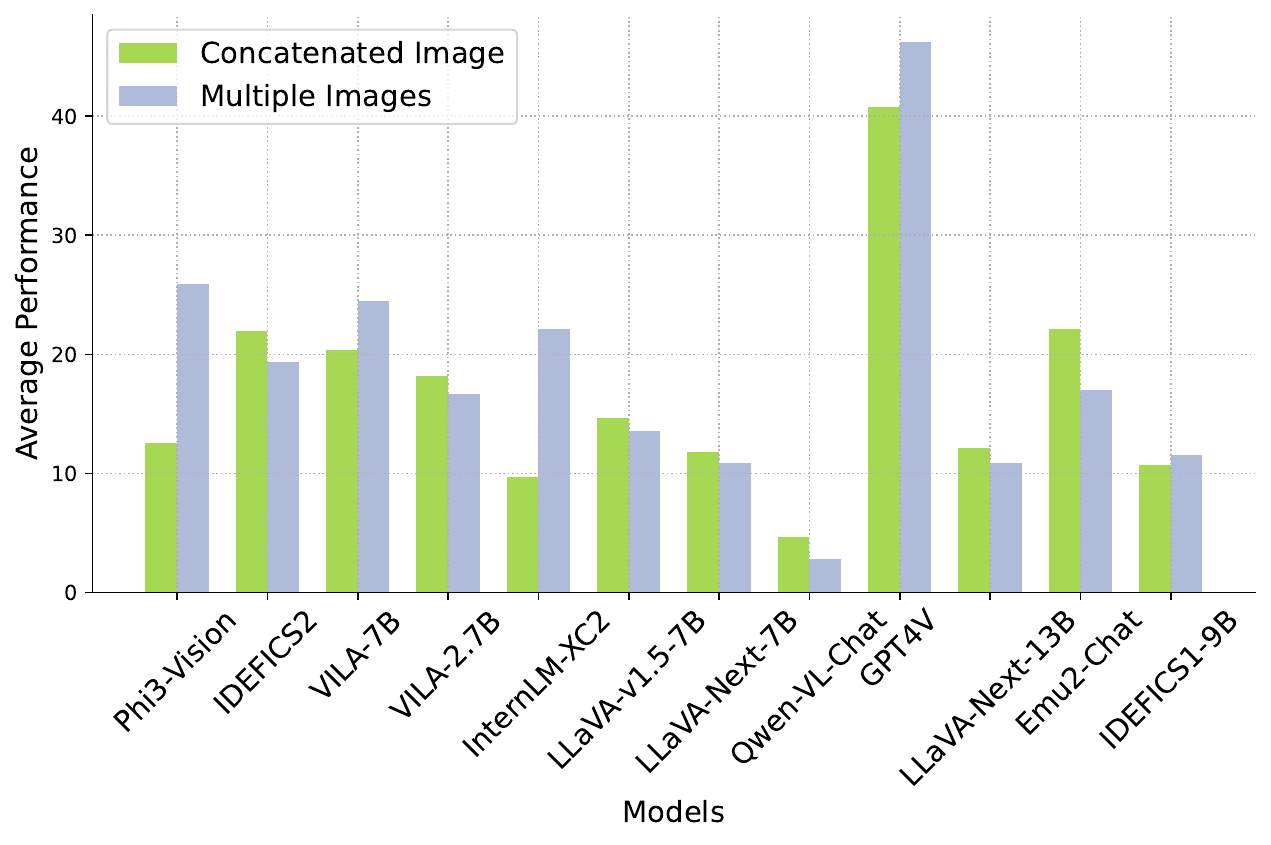}
    \caption{Comparison of image input format: Multi-image vs single concatenated image.}
    \label{fig:concat}
    \end{subfigure}
    \\
    \begin{subfigure}[t]{0.49\textwidth}
    \includegraphics[width=\linewidth]{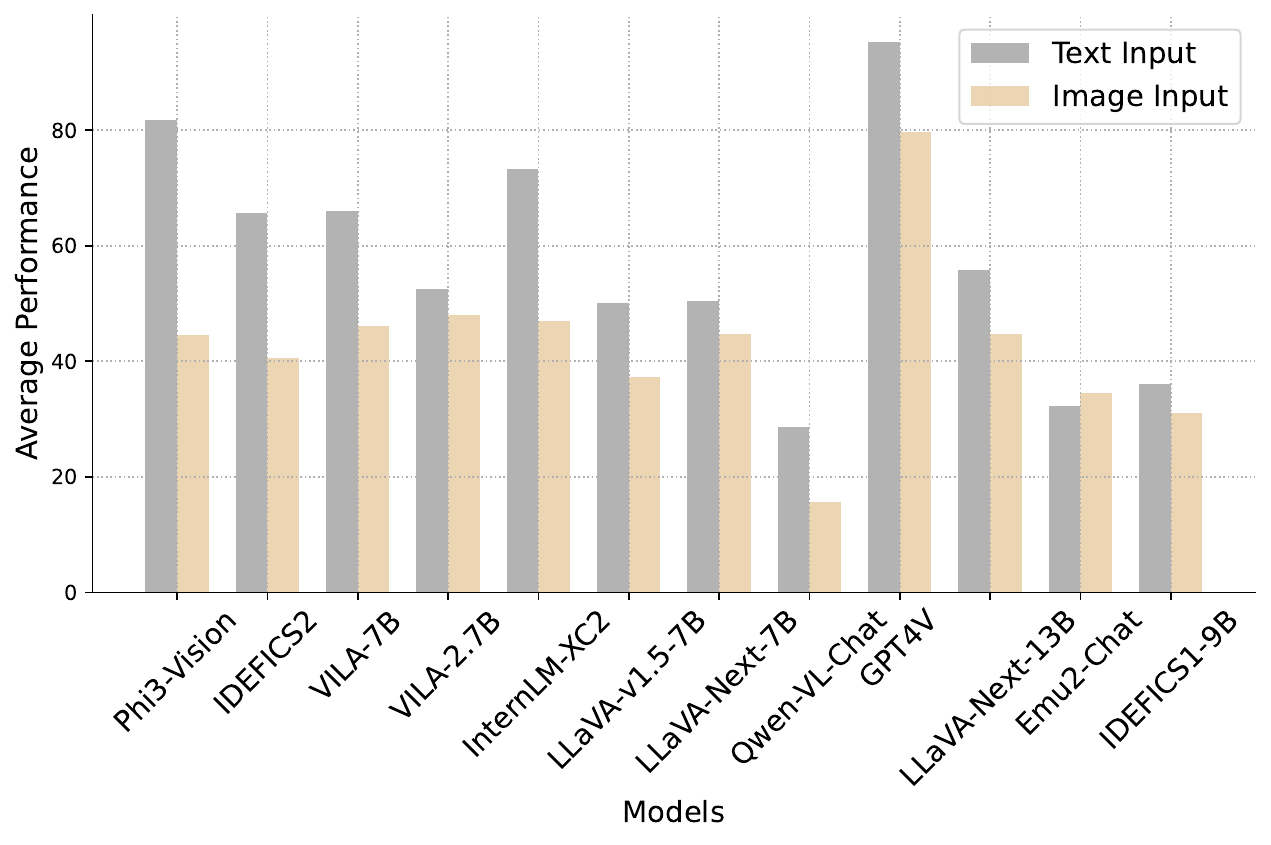}
    \caption{Comparison of text and image inputs. Inputting equivalent text is usually better.}
    \label{fig:text}
    \end{subfigure}
    \hfill
    \begin{subfigure}[t]{0.49\textwidth}
    \includegraphics[width=\linewidth]{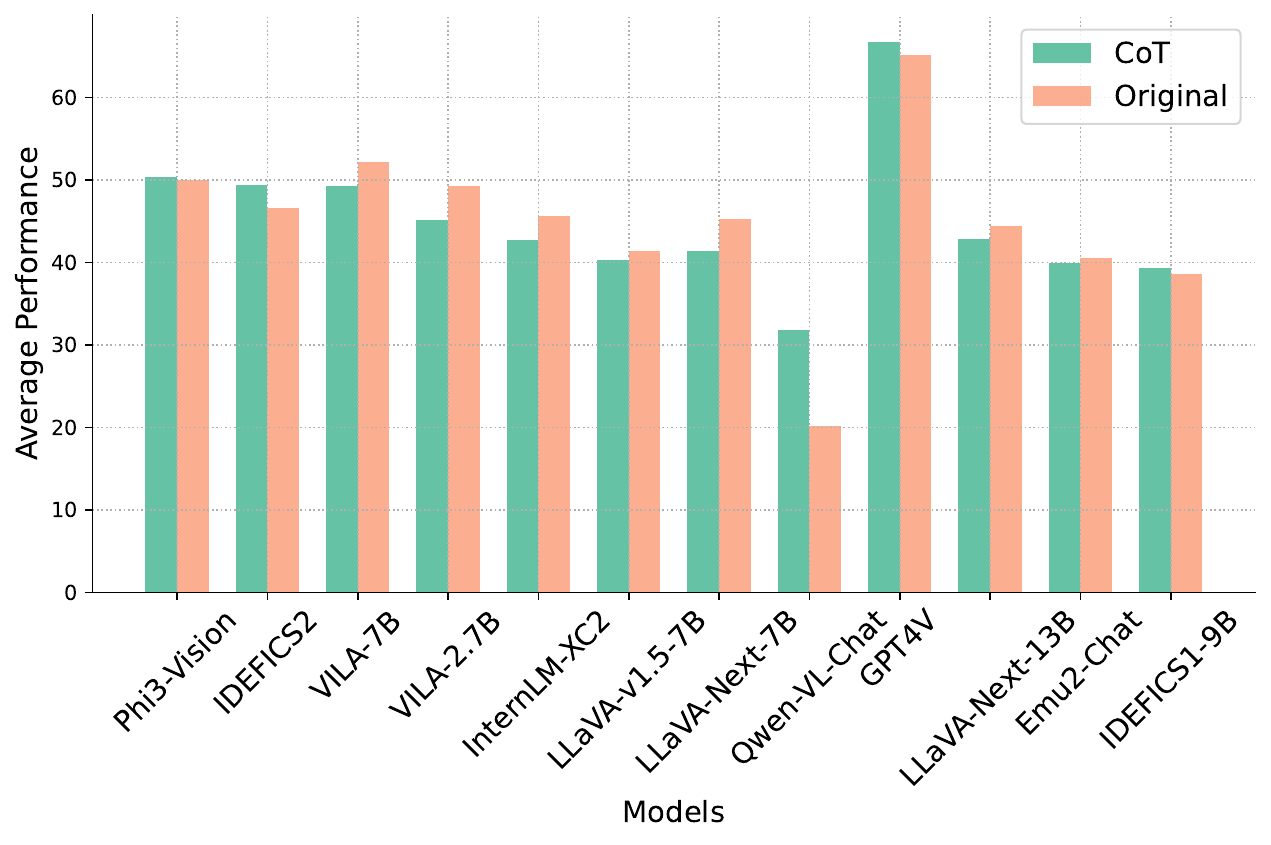}
    \caption{Comparison between original task and Chain-of-Thought (CoT) prompting. CoT is not helpful.}
    \label{fig:CoT}
    \end{subfigure}
    \caption{Performance analysis on \bench.}
    \label{fig:analysis}
\end{figure}

Prior VLM studies have debated how to encode multiple image inputs \citep{liu2024llavanext}. One hypothesis is that image separator challenges can be avoided by concatenating the original images into a single larger image. We test this alternative encoding for subsets including 3D scene understanding, counting, and synthetic visual logic chain. The concatenated image is composed of $1 \times 2$ or $2 \times 2$ original images, with red lines denoting the boundary. We fill in the vacancy with a pure white picture in case only 3 images are available. The results in Figure~\ref{fig:concat} show that most models are worse with this encoding of concatenating images.
One notable exception is the LLaVA model family, where the model is trained on data with only one image as input, the performance on concatenated images improves due to the reduced shift in the number of input images between training and testing.

\subsection{LVLMs Understand Text Inputs Better than Images}
For the tasks involving text understanding, we consider a non-end-to-end baseline of translating images to text, and then inputting the result directly to the LLM without using a vision encoder. 
Specifically, for code understanding and plot code understanding subsets, we use the text source code as inputs. 
For Arxiv paper citation look up subset, we use existing OCR tools PdfReader\footnote{https://github.com/maxpmaxp/pdfreader} to extract the text content from the papers.
The results in Figure~\ref{fig:text} show that text inputs are usually better than image inputs, and sometimes substantially better. 
However, it is hard to disentangle how much this effect is due to reduced difficulty of perception, versus tending to induce a smaller number of tokens and thus putting less stress on the underlying LLMs' long-context abilities.

\subsection{Chain-of-Thought Does Not Help Reasoning Over Multiple Images}
We utilize zero-shot chain-of-thought prompting~\citep{wei2022chain, kojima2022large} for several subsets that require complicated reasoning, including visual analogy, attribute matching, code understanding, plot code understanding, and food comparisons. Figure~\ref{fig:CoT} illustrates the comparisons between CoT and the original prompting. The results show only minor improvements or sometimes even decreases with CoT prompting, indicating that simple prompting techniques cannot effectively address the reasoning of multiple images but require fundamental improvements during training.

\section{Conclusion and Discussion}
In this paper, we introduced \bench, a benchmark specifically designed to evaluate the multi-image reasoning capabilities of visual language models (LVLMs). Our benchmark covers four categories: visual world knowledge, multi-image reasoning, perception, and multi-hop reasoning, addressing a significant gap in existing evaluations for LVLMs that focus predominantly on single-image tasks.
Our comprehensive evaluation revealed a notable performance gap in these reasoning tasks. 
While open-source models like LLaVA perform well on single-image tasks, they lag significantly behind in multi-image scenarios compared to the close-sourced GPT4-V. 
Most interestingly, on the dimension of visual world knowledge and multi-hop reasoning scenarios, we did not observe any open-sourced model that is able to consistently outperform the random chance baseline.
This highlights the inherent complexity of multi-image reasoning and the need for further advancements in this area.

We have also conducted a set of analysis experiments to provide more insights on multi-image evaluation. 
We demonstrated that the questions with \bench~cannot be answered by only utilizing any single images, and they require reasoning across multiple images.
The method of incorporating multiple images in the model input has also been discussed, we compared the method that uses one large concatenated image as one single image input and the method that encodes the multiple images separately. The result shows that simply concatenating multiple images together do not yield a good performance.
In tasks that require the model to recognize the text on the image, we perform experiments of first extracting the text and then performing reasoning using pure language. The results show that these LVLMs can reason better using pure text than multiple images as the input. Indicating a deficiency of these LVLMs -- they cannot reliably decode the text presented in the image.
Lastly, we tested one simple prompting method -- chain-of-thought prompting -- that have already demonstrated its effectiveness for language reasoning on \bench. Yet despite the effectiveness on language tasks, it did not demonstrate any notable improvements.

We believe that \bench~underscores the necessity for enhanced model architectures and training datasets that can better handle multi-image contexts. 
And that \bench~will serve as a valuable resource for the research community, facilitating the advancement of VLMs and enabling more complex and nuanced applications across various domains.
In summary, \bench~not only highlights the current limitations in multi-image reasoning but also provides a clear roadmap for future research, paving the way for more capable and intelligent multi-modal models.

\keypoint{Limitations and Broader Impact.}
\bench~aims to provide a platform for evaluating the reasoning ability of LVLMs given multiple image inputs. Like all benchmark works, while we have considered as many scenarios for evaluation as possible, \bench~cannot reflect every possible real-world use case. For critical scenarios such as self-driving cars, additional attention on how to test the safety of the system is required.
As \bench~contains general images we carefully curated to test the reasoning abilities of LVLMs, we do not anticipate any possible major negative impact to the society.

\bibliography{ref}

\begin{thebibliography}{49}
\providecommand{\natexlab}[1]{#1}
\providecommand{\url}[1]{\texttt{#1}}
\expandafter\ifx\csname urlstyle\endcsname\relax
  \providecommand{\doi}[1]{doi: #1}\else
  \providecommand{\doi}{doi: \begingroup \urlstyle{rm}\Url}\fi

\bibitem[Abdin et~al.(2024)Abdin, Jacobs, Awan, Aneja, Awadallah, Awadalla, Bach, Bahree, Bakhtiari, Behl, et~al.]{abdin2024phi3}
Marah Abdin, Sam~Ade Jacobs, Ammar~Ahmad Awan, Jyoti Aneja, Ahmed Awadallah, Hany Awadalla, Nguyen Bach, Amit Bahree, Arash Bakhtiari, Harkirat Behl, et~al.
\newblock Phi-3 technical report: A highly capable language model locally on your phone.
\newblock \emph{arXiv preprint arXiv:2404.14219}, 2024.

\bibitem[Antol et~al.(2015)Antol, Agrawal, Lu, Mitchell, Batra, Zitnick, and Parikh]{VQA}
Stanislaw Antol, Aishwarya Agrawal, Jiasen Lu, Margaret Mitchell, Dhruv Batra, C.~Lawrence Zitnick, and Devi Parikh.
\newblock {VQA}: {V}isual {Q}uestion {A}nswering.
\newblock In \emph{International Conference on Computer Vision (ICCV)}, 2015.

\bibitem[Bai et~al.(2023)Bai, Bai, Yang, Wang, Tan, Wang, Lin, Zhou, and Zhou]{bai2023qwen}
Jinze Bai, Shuai Bai, Shusheng Yang, Shijie Wang, Sinan Tan, Peng Wang, Junyang Lin, Chang Zhou, and Jingren Zhou.
\newblock Qwen-vl: A frontier large vision-language model with versatile abilities.
\newblock \emph{arXiv preprint arXiv:2308.12966}, 2023.

\bibitem[Bitton et~al.(2023)Bitton, Yosef, Strugo, Shahaf, Schwartz, and Stanovsky]{bitton2023vasr}
Yonatan Bitton, Ron Yosef, Eliyahu Strugo, Dafna Shahaf, Roy Schwartz, and Gabriel Stanovsky.
\newblock Vasr: Visual analogies of situation recognition.
\newblock In \emph{Proceedings of the AAAI Conference on Artificial Intelligence}, 2023.

\bibitem[Chen et~al.(2023)Chen, Zhu, Shen, Li, Liu, Zhang, Krishnamoorthi, Chandra, Xiong, and Elhoseiny]{chen2023minigpt}
Jun Chen, Deyao Zhu, Xiaoqian Shen, Xiang Li, Zechun Liu, Pengchuan Zhang, Raghuraman Krishnamoorthi, Vikas Chandra, Yunyang Xiong, and Mohamed Elhoseiny.
\newblock Minigpt-v2: large language model as a unified interface for vision-language multi-task learning.
\newblock \emph{arXiv preprint arXiv:2310.09478}, 2023.

\bibitem[Chen et~al.(2021)Chen, Tworek, Jun, Yuan, Pinto, Kaplan, Edwards, Burda, Joseph, Brockman, et~al.]{humaneval}
Mark Chen, Jerry Tworek, Heewoo Jun, Qiming Yuan, Henrique Ponde de~Oliveira Pinto, Jared Kaplan, Harri Edwards, Yuri Burda, Nicholas Joseph, Greg Brockman, et~al.
\newblock Evaluating large language models trained on code.
\newblock \emph{arXiv preprint arXiv:2107.03374}, 2021.

\bibitem[Contributors(2023)]{autogpt}
AutoGPT Contributors.
\newblock Autogpt.
\newblock \emph{https://news.agpt.co/}, 2023.

\bibitem[Dai et~al.(2023)Dai, Li, Li, Tiong, Zhao, Wang, Li, Fung, and Hoi]{dai2023instructblip}
Wenliang Dai, Junnan Li, Dongxu Li, Anthony Tiong, Junqi Zhao, Weisheng Wang, Boyang Li, Pascale Fung, and Steven Hoi.
\newblock Instruct{BLIP}: Towards general-purpose vision-language models with instruction tuning.
\newblock In \emph{NeurIPS}, 2023.

\bibitem[Dong et~al.(2024)Dong, Zhang, Zang, Cao, Wang, Ouyang, Wei, Zhang, Duan, Cao, et~al.]{dong2024internlmx2}
Xiaoyi Dong, Pan Zhang, Yuhang Zang, Yuhang Cao, Bin Wang, Linke Ouyang, Xilin Wei, Songyang Zhang, Haodong Duan, Maosong Cao, et~al.
\newblock Internlm-xcomposer2: Mastering free-form text-image composition and comprehension in vision-language large model.
\newblock \emph{arXiv preprint arXiv:2401.16420}, 2024.

\bibitem[Fu et~al.(2023)Fu, Chen, Shen, Qin, Zhang, Lin, Yang, Zheng, Li, Sun, et~al.]{fu2023mme}
Chaoyou Fu, Peixian Chen, Yunhang Shen, Yulei Qin, Mengdan Zhang, Xu~Lin, Jinrui Yang, Xiawu Zheng, Ke~Li, Xing Sun, et~al.
\newblock Mme: A comprehensive evaluation benchmark for multimodal large language models.
\newblock \emph{arXiv preprint arXiv:2306.13394}, 2023.

\bibitem[Fu et~al.(2024)Fu, Hu, Li, Feng, Wang, Lin, Roth, Smith, Ma, and Krishna]{fu2024blink}
Xingyu Fu, Yushi Hu, Bangzheng Li, Yu~Feng, Haoyu Wang, Xudong Lin, Dan Roth, Noah~A Smith, Wei-Chiu Ma, and Ranjay Krishna.
\newblock Blink: Multimodal large language models can see but not perceive.
\newblock \emph{arXiv preprint arXiv:2404.12390}, 2024.

\bibitem[Goyal et~al.(2017)Goyal, Khot, Summers-Stay, Batra, and Parikh]{goyal2017vqaMatter}
Y.~Goyal, T.~Khot, D.~Summers-Stay, D.~Batra, and D.~Parikh.
\newblock Making the v in vqa matter: Elevating the role of image understanding in visual question answering.
\newblock In \emph{CVPR}, 2017.

\bibitem[Gurari et~al.(2018)Gurari, Li, Stangl, Guo, Lin, Grauman, Luo, and Bigham]{gurari2018vizwiz}
Danna Gurari, Qing Li, Abigale~J Stangl, Anhong Guo, Chi Lin, Kristen Grauman, Jiebo Luo, and Jeffrey~P Bigham.
\newblock Vizwiz grand challenge: Answering visual questions from blind people.
\newblock In \emph{CVPR}, pages 3608--3617, 2018.

\bibitem[Hendrycks et~al.(2020)Hendrycks, Burns, Basart, Zou, Mazeika, Song, and Steinhardt]{mmlu}
Dan Hendrycks, Collin Burns, Steven Basart, Andy Zou, Mantas Mazeika, Dawn Song, and Jacob Steinhardt.
\newblock Measuring massive multitask language understanding.
\newblock \emph{arXiv preprint arXiv:2009.03300}, 2020.

\bibitem[Hendrycks et~al.(2021{\natexlab{a}})Hendrycks, Basart, Mu, Kadavath, Wang, Dorundo, Desai, Zhu, Parajuli, Guo, et~al.]{hendrycks2020many}
Dan Hendrycks, Steven Basart, Norman Mu, Saurav Kadavath, Frank Wang, Evan Dorundo, Rahul Desai, Tyler~Lixuan Zhu, Samyak Parajuli, Mike Guo, et~al.
\newblock The many faces of robustness: A critical analysis of out-of-distribution generalization. 2021 ieee.
\newblock In \emph{CVF International Conference on Computer Vision (ICCV)}, 2021{\natexlab{a}}.

\bibitem[Hendrycks et~al.(2021{\natexlab{b}})Hendrycks, Burns, Kadavath, Arora, Basart, Tang, Song, and Steinhardt]{math}
Dan Hendrycks, Collin Burns, Saurav Kadavath, Akul Arora, Steven Basart, Eric Tang, Dawn Song, and Jacob Steinhardt.
\newblock Measuring mathematical problem solving with the math dataset.
\newblock \emph{arXiv preprint arXiv:2103.03874}, 2021{\natexlab{b}}.

\bibitem[Kojima et~al.(2022)Kojima, Gu, Reid, Matsuo, and Iwasawa]{kojima2022large}
Takeshi Kojima, Shixiang~Shane Gu, Machel Reid, Yutaka Matsuo, and Yusuke Iwasawa.
\newblock Large language models are zero-shot reasoners.
\newblock \emph{Advances in neural information processing systems}, 2022.

\bibitem[Lauren{\c{c}}on et~al.(2023)Lauren{\c{c}}on, Saulnier, Tronchon, Bekman, Singh, Lozhkov, Wang, Karamcheti, Rush, Kiela, et~al.]{laurenccon2024obelics}
Hugo Lauren{\c{c}}on, Lucile Saulnier, L{\'e}o Tronchon, Stas Bekman, Amanpreet Singh, Anton Lozhkov, Thomas Wang, Siddharth Karamcheti, Alexander Rush, Douwe Kiela, et~al.
\newblock Obelics: An open web-scale filtered dataset of interleaved image-text documents.
\newblock \emph{NeurIPS}, 36, 2023.

\bibitem[Lauren{\c{c}}on et~al.(2024)Lauren{\c{c}}on, Tronchon, Cord, and Sanh]{laurenccon2024idefics2}
Hugo Lauren{\c{c}}on, L{\'e}o Tronchon, Matthieu Cord, and Victor Sanh.
\newblock What matters when building vision-language models?
\newblock \emph{arXiv preprint arXiv:2405.02246}, 2024.

\bibitem[Li et~al.(2023{\natexlab{a}})Li, Ge, Ge, Wang, Wang, Zhang, and Shan]{li2023Seed-bench-2}
Bohao Li, Yuying Ge, Yixiao Ge, Guangzhi Wang, Rui Wang, Ruimao Zhang, and Ying Shan.
\newblock Seed-bench-2: Benchmarking multimodal large language models.
\newblock \emph{arXiv preprint arXiv:2311.17092}, 2023{\natexlab{a}}.

\bibitem[Li et~al.(2023{\natexlab{b}})Li, Wang, Wang, Ge, Ge, and Shan]{li2023seedbench}
Bohao Li, Rui Wang, Guangzhi Wang, Yuying Ge, Yixiao Ge, and Ying Shan.
\newblock Seed-bench: Benchmarking multimodal llms with generative comprehension.
\newblock \emph{arXiv preprint arXiv:2307.16125}, 2023{\natexlab{b}}.

\bibitem[Li et~al.(2024)Li, Pan, Ge, Gao, Ji, Zhang, Chua, Tang, Zhang, and Zhuang]{li2023fine}
Juncheng Li, Kaihang Pan, Zhiqi Ge, Minghe Gao, Wei Ji, Wenqiao Zhang, Tat-Seng Chua, Siliang Tang, Hanwang Zhang, and Yueting Zhuang.
\newblock Fine-tuning multimodal llms to follow zero-shot demonstrative instructions.
\newblock In \emph{ICLR}, 2024.

\bibitem[Lin et~al.(2024)Lin, Yin, Ping, Lu, Molchanov, Tao, Mao, Kautz, Shoeybi, and Han]{lin2024vila}
Ji~Lin, Hongxu Yin, Wei Ping, Yao Lu, Pavlo Molchanov, Andrew Tao, Huizi Mao, Jan Kautz, Mohammad Shoeybi, and Song Han.
\newblock Vila: On pre-training for visual language models.
\newblock \emph{CVPR}, 2024.

\bibitem[Lin et~al.(2014)Lin, Maire, Belongie, Hays, Perona, Ramanan, Doll{\'a}r, and Zitnick]{lin2014microsoft}
Tsung-Yi Lin, Michael Maire, Serge Belongie, James Hays, Pietro Perona, Deva Ramanan, Piotr Doll{\'a}r, and C~Lawrence Zitnick.
\newblock Microsoft coco: Common objects in context.
\newblock In \emph{ECCV}, 2014.

\bibitem[Liu et~al.(2023{\natexlab{a}})Liu, Li, Li, and Lee]{liu2023improved}
Haotian Liu, Chunyuan Li, Yuheng Li, and Yong~Jae Lee.
\newblock Improved baselines with visual instruction tuning.
\newblock \emph{arXiv preprint arXiv:2310.03744}, 2023{\natexlab{a}}.

\bibitem[Liu et~al.(2023{\natexlab{b}})Liu, Li, Wu, and Lee]{liu2023visual}
Haotian Liu, Chunyuan Li, Qingyang Wu, and Yong~Jae Lee.
\newblock Visual instruction tuning.
\newblock \emph{NeurIPS}, 2023{\natexlab{b}}.

\bibitem[Liu et~al.(2024)Liu, Li, Li, Li, Zhang, Shen, and Lee]{liu2024llavanext}
Haotian Liu, Chunyuan Li, Yuheng Li, Bo~Li, Yuanhan Zhang, Sheng Shen, and Yong~Jae Lee.
\newblock Llava-next: Improved reasoning, ocr, and world knowledge, January 2024.
\newblock URL \url{https://llava-vl.github.io/blog/2024-01-30-llava-next/}.

\bibitem[Liu et~al.(2023{\natexlab{c}})Liu, Duan, Zhang, Li, Zhang, Zhao, Yuan, Wang, He, Liu, et~al.]{liu2023mmbench}
Yuan Liu, Haodong Duan, Yuanhan Zhang, Bo~Li, Songyang Zhang, Wangbo Zhao, Yike Yuan, Jiaqi Wang, Conghui He, Ziwei Liu, et~al.
\newblock Mmbench: Is your multi-modal model an all-around player?
\newblock \emph{arXiv preprint arXiv:2307.06281}, 2023{\natexlab{c}}.

\bibitem[Lu et~al.(2022)Lu, Mishra, Xia, Qiu, Chang, Zhu, Tafjord, Clark, and Kalyan]{lu2022scienceqa}
Pan Lu, Swaroop Mishra, Tony Xia, Liang Qiu, Kai-Wei Chang, Song-Chun Zhu, Oyvind Tafjord, Peter Clark, and Ashwin Kalyan.
\newblock Learn to explain: Multimodal reasoning via thought chains for science question answering.
\newblock In \emph{NeurIPS}, 2022.

\bibitem[Lu et~al.(2024)Lu, Bansal, Xia, Liu, Li, Hajishirzi, Cheng, Chang, Galley, and Gao]{lu2024mathvista}
Pan Lu, Hritik Bansal, Tony Xia, Jiacheng Liu, Chunyuan Li, Hannaneh Hajishirzi, Hao Cheng, Kai-Wei Chang, Michel Galley, and Jianfeng Gao.
\newblock Mathvista: Evaluating mathematical reasoning of foundation models in visual contexts.
\newblock In \emph{International Conference on Learning Representations (ICLR)}, 2024.

\bibitem[Mishra et~al.(2019)Mishra, Shekhar, Singh, and Chakraborty]{ocrvqa}
Anand Mishra, Shashank Shekhar, Ajeet~Kumar Singh, and Anirban Chakraborty.
\newblock Ocr-vqa: Visual question answering by reading text in images.
\newblock In \emph{ICDAR}, 2019.

\bibitem[OpenAI(2023)]{openai2023gpt4}
OpenAI.
\newblock Gpt-4 technical report.
\newblock \emph{arXiv}, 2023.

\bibitem[OpenFoodFact(2024)]{openfoodfact}
OpenFoodFact.
\newblock Openfoodfact database.
\newblock \emph{\url{https://world.openfoodfacts.org/}}, 2024.

\bibitem[Reid et~al.(2024)Reid, Savinov, Teplyashin, Lepikhin, Lillicrap, Alayrac, Soricut, Lazaridou, Firat, Schrittwieser, et~al.]{reid2024gemini}
Machel Reid, Nikolay Savinov, Denis Teplyashin, Dmitry Lepikhin, Timothy Lillicrap, Jean-baptiste Alayrac, Radu Soricut, Angeliki Lazaridou, Orhan Firat, Julian Schrittwieser, et~al.
\newblock Gemini 1.5: Unlocking multimodal understanding across millions of tokens of context.
\newblock \emph{arXiv preprint arXiv:2403.05530}, 2024.

\bibitem[Shang et~al.(2024)Shang, Cai, Xu, Lee, and Yan]{shang2024llava}
Yuzhang Shang, Mu~Cai, Bingxin Xu, Yong~Jae Lee, and Yan Yan.
\newblock Llava-prumerge: Adaptive token reduction for efficient large multimodal models.
\newblock \emph{arXiv preprint arXiv:2403.15388}, 2024.

\bibitem[Singh et~al.(2021)Singh, Pang, Toh, Huang, Galuba, and Hassner]{singh2021textocr}
Amanpreet Singh, Guan Pang, Mandy Toh, Jing Huang, Wojciech Galuba, and Tal Hassner.
\newblock {TextOCR}: Towards large-scale end-to-end reasoning for arbitrary-shaped scene text.
\newblock In \emph{CVPR}, 2021.

\bibitem[Sun et~al.(2023)Sun, Cui, Zhang, Zhang, Yu, Luo, Wang, Rao, Liu, Huang, et~al.]{sun2023emu2}
Quan Sun, Yufeng Cui, Xiaosong Zhang, Fan Zhang, Qiying Yu, Zhengxiong Luo, Yueze Wang, Yongming Rao, Jingjing Liu, Tiejun Huang, et~al.
\newblock Generative multimodal models are in-context learners.
\newblock \emph{arXiv preprint arXiv:2312.13286}, 2023.

\bibitem[Touvron et~al.(2023)Touvron, Martin, Stone, Albert, Almahairi, Babaei, Bashlykov, Batra, Bhargava, Bhosale, et~al.]{touvron2023llama}
Hugo Touvron, Louis Martin, Kevin Stone, Peter Albert, Amjad Almahairi, Yasmine Babaei, Nikolay Bashlykov, Soumya Batra, Prajjwal Bhargava, Shruti Bhosale, et~al.
\newblock Llama 2: Open foundation and fine-tuned chat models.
\newblock \emph{arXiv preprint arXiv:2307.09288}, 2023.

\bibitem[Tu et~al.(2023)Tu, Cui, Wang, Zhou, Zhao, Han, Zhou, Yao, and Xie]{tu2023how}
Haoqin Tu, Chenhang Cui, Zijun Wang, Yiyang Zhou, Bingchen Zhao, Junlin Han, Wangchunshu Zhou, Huaxiu Yao, and Cihang Xie.
\newblock How many unicorns are in this image? a safety evaluation benchmark for vision llms.
\newblock \emph{arXiv preprint arXiv:2311.16101}, 2023.

\bibitem[Wang et~al.(2024)Wang, Zhou, Liu, Lu, Xu, He, Yoon, Lu, Bertasius, Bansal, et~al.]{wang2024mementos}
Xiyao Wang, Yuhang Zhou, Xiaoyu Liu, Hongjin Lu, Yuancheng Xu, Feihong He, Jaehong Yoon, Taixi Lu, Gedas Bertasius, Mohit Bansal, et~al.
\newblock Mementos: A comprehensive benchmark for multimodal large language model reasoning over image sequences.
\newblock \emph{arXiv preprint arXiv:2401.10529}, 2024.

\bibitem[Wei et~al.(2022)Wei, Wang, Schuurmans, Bosma, Xia, Chi, Le, Zhou, et~al.]{wei2022chain}
Jason Wei, Xuezhi Wang, Dale Schuurmans, Maarten Bosma, Fei Xia, Ed~Chi, Quoc~V Le, Denny Zhou, et~al.
\newblock Chain-of-thought prompting elicits reasoning in large language models.
\newblock \emph{NeurIPS}, 2022.

\bibitem[Wu et~al.(2023)Wu, Zhang, Zhang, Chen, Liao, Wang, Li, Sun, Yan, Zhai, et~al.]{wu2023q}
Haoning Wu, Zicheng Zhang, Erli Zhang, Chaofeng Chen, Liang Liao, Annan Wang, Chunyi Li, Wenxiu Sun, Qiong Yan, Guangtao Zhai, et~al.
\newblock Q-bench: A benchmark for general-purpose foundation models on low-level vision.
\newblock \emph{arXiv preprint arXiv:2309.14181}, 2023.

\bibitem[Yu et~al.(2023)Yu, Yang, Li, Wang, Lin, Liu, Wang, and Wang]{yu2023mmvet}
Weihao Yu, Zhengyuan Yang, Linjie Li, Jianfeng Wang, Kevin Lin, Zicheng Liu, Xinchao Wang, and Lijuan Wang.
\newblock Mm-vet: Evaluating large multimodal models for integrated capabilities.
\newblock \emph{arXiv preprint arXiv:2308.02490}, 2023.

\bibitem[Yue et~al.(2024)Yue, Ni, Zhang, Zheng, Liu, Zhang, Stevens, Jiang, Ren, Sun, Wei, Yu, Yuan, Sun, Yin, Zheng, Yang, Liu, Huang, Sun, Su, and Chen]{yue2023mmmu}
Xiang Yue, Yuansheng Ni, Kai Zhang, Tianyu Zheng, Ruoqi Liu, Ge~Zhang, Samuel Stevens, Dongfu Jiang, Weiming Ren, Yuxuan Sun, Cong Wei, Botao Yu, Ruibin Yuan, Renliang Sun, Ming Yin, Boyuan Zheng, Zhenzhu Yang, Yibo Liu, Wenhao Huang, Huan Sun, Yu~Su, and Wenhu Chen.
\newblock Mmmu: A massive multi-discipline multimodal understanding and reasoning benchmark for expert agi.
\newblock In \emph{Proceedings of CVPR}, 2024.

\bibitem[Zhang et~al.(2024)Zhang, Zhai, Zhao, Zong, Wen, and Zhao]{Zhang_2024_CVPR}
Letian Zhang, Xiaotong Zhai, Zhongkai Zhao, Yongshuo Zong, Xin Wen, and Bingchen Zhao.
\newblock What if the tv was off? examining counterfactual reasoning abilities of multi-modal language models.
\newblock In \emph{CVPR}, 2024.

\bibitem[Zhao et~al.(2023{\natexlab{a}})Zhao, Cui, Wu, Yoshie, Yang, and Aodha]{zhao23mug}
Bingchen Zhao, Quan Cui, Hao Wu, Osamu Yoshie, Cheng Yang, and Oisin~Mac Aodha.
\newblock Vision learners meet web image-text pairs.
\newblock \emph{arXiv preprint arXiv:2301.07088}, 2023{\natexlab{a}}.

\bibitem[Zhao et~al.(2023{\natexlab{b}})Zhao, Tu, Wei, Mei, and Xie]{zhao2023tuning}
Bingchen Zhao, Haoqin Tu, Chen Wei, Jieru Mei, and Cihang Xie.
\newblock Tuning layernorm in attention: Towards efficient multi-modal llm finetuning.
\newblock \emph{arXiv preprint arXiv:2312.11420}, 2023{\natexlab{b}}.

\bibitem[Zhu et~al.(2023)Zhu, Hessel, Awadalla, Gadre, Dodge, Fang, Yu, Schmidt, Wang, and Choi]{zhu2023multimodal}
Wanrong Zhu, Jack Hessel, Anas Awadalla, Samir~Yitzhak Gadre, Jesse Dodge, Alex Fang, Youngjae Yu, Ludwig Schmidt, William~Yang Wang, and Yejin Choi.
\newblock {Multimodal C4}: An open, billion-scale corpus of images interleaved with text.
\newblock \emph{arXiv preprint arXiv:2304.06939}, 2023.

\bibitem[Zong et~al.(2023)Zong, Yu, Zhao, Chavhan, and Hospedales]{zong2023fool}
Yongshuo Zong, Tingyang Yu, Bingchen Zhao, Ruchika Chavhan, and Timothy Hospedales.
\newblock Fool your (vision and) language model with embarrassingly simple permutations.
\newblock \emph{arXiv preprint arXiv:2310.01651}, 2023.

\end{thebibliography}
\bibliographystyle{plainnat}

\clearpage
\section*{Checklist}

\begin{enumerate}

\item For all authors...
\begin{enumerate}
  \item Do the main claims made in the abstract and introduction accurately reflect the paper's contributions and scope?
    \answerYes{}
  \item Did you describe the limitations of your work?
    \answerYes{}
  \item Did you discuss any potential negative societal impacts of your work?
    \answerYes{}
  \item Have you read the ethics review guidelines and ensured that your paper conforms to them?
    \answerYes{}
\end{enumerate}

\item If you are including theoretical results...
\begin{enumerate}
  \item Did you state the full set of assumptions of all theoretical results?
    \answerNA{}
	\item Did you include complete proofs of all theoretical results?
     \answerNA{}
\end{enumerate}

\item If you ran experiments (e.g. for benchmarks)...
\begin{enumerate}
  \item Did you include the code, data, and instructions needed to reproduce the main experimental results (either in the supplemental material or as a URL)?
    \answerYes{We will include these in the supplementary materials.}
  \item Did you specify all the training details (e.g., data splits, hyperparameters, how they were chosen)?
    \answerNA{We did not train any new models.}
	\item Did you report error bars (e.g., with respect to the random seed after running experiments multiple times)?
    \answerNA{Our evaluation uses fixed random seeds and is fully reproducible}
	\item Did you include the total amount of compute and the type of resources used (e.g., type of GPUs, internal cluster, or cloud provider)?
    \answerYes{We will include this in the supplementary.}
\end{enumerate}

\item If you are using existing assets (e.g., code, data, models) or curating/releasing new assets...
\begin{enumerate}
  \item If your work uses existing assets, did you cite the creators?
    \answerYes{}
  \item Did you mention the license of the assets?
    \answerYes{}
  \item Did you include any new assets either in the supplemental material or as a URL?
    \answerYes{}
  \item Did you discuss whether and how consent was obtained from people whose data you're using/curating?
    \answerYes{}
  \item Did you discuss whether the data you are using/curating contains personally identifiable information or offensive content?
    \answerYes{}
\end{enumerate}

\item If you used crowdsourcing or conducted research with human subjects...
\begin{enumerate}
  \item Did you include the full text of instructions given to participants and screenshots, if applicable?
     \answerNA{}
  \item Did you describe any potential participant risks, with links to Institutional Review Board (IRB) approvals, if applicable?
     \answerNA{}
  \item Did you include the estimated hourly wage paid to participants and the total amount spent on participant compensation?
    \answerNA{}
\end{enumerate}

\end{enumerate}

\clearpage
\appendix
\appendix

\section{Dataset Access and Copyright Statement}

\bench~can be accessed on HuggingFace~\footnote{\url{https://huggingface.co/datasets/VLLMs/MIRB}}. Our complete code for evaluating models can be accessed on GitHub~\footnote{\url{https://github.com/DTennant/MIRB_eval}}.
The documentations of the dataset and the usage of the evaluation code is available on the GitHub.
\bench~is intended for the sole purpose of benchmarking the reasoning abilities of visual language models on multiple images.
The benchmark score on \bench~onlt reflects the abilities in understanding and reasoning through multiple images and thus cannot be used as the only metric to determine the capability of visual language models.
All images in~\bench~are collected and filtered by the author, and the authors take full responsibility in case of violation of copyrights of these images.

\begin{table}[h]
    \centering
    \begin{tabular}{lcc}
    \toprule
    Dimension     &  Data Source & \# Imgs\\
    \midrule
    Code understanding     &  Popular Python Libraries   &  141   \\
    Plot understanding     &  Matplotlib Tutorials  &   352  \\
    Visual analogy         &  \citet{bitton2023vasr}  &  493   \\
    3D scene understanding &  Blender Rendering  &  270   \\
    Sightseeing            &  Pixabay  &  105   \\
    Food comparisons       &  OpenFoodFact~\citep{openfoodfact}  &   222  \\
    Image jigsaw           &  Pixsum  &  500   \\
    Counting               &  COCO~\citep{lin2014microsoft}  &  71   \\
    Attribute matching     &  ImageNet-R~\citep{hendrycks2020many}  &  400   \\
    Synthetic logic chain  &  Pixsum  & 186    \\
    ArXiv paper citation   &  ArXiv  & 487    \\
    \bottomrule
    \end{tabular}
    \caption{Detailed list of data sources.}
    \label{tab:data_source}
\end{table}

\section{Data Source}
Table~\ref{tab:data_source} provides a detailed list of the data sources we use to collect the images.
For the images we use to evaluate the dimension of code understanding and plot understanding, we gather code snippets from popular Python libraries and render them as images using Playwright with syntax highlighting.
The libraries include, \texttt{requests}\footnote{https://pypi.org/project/requests/}, \texttt{flask}\footnote{https://flask.palletsprojects.com/en/3.0.x/}, \texttt{youtube-dl}\footnote{https://github.com/ytdl-org/youtube-dl}, \texttt{pandas}\footnote{https://pandas.pydata.org/}, \texttt{pygame}\footnote{https://www.pygame.org/}, \texttt{beautifulsoup}\footnote{https://beautiful-soup-4.readthedocs.io/en/latest/}, \texttt{matplotlib}\footnote{https://matplotlib.org/}, and \texttt{numpy}\footnote{https://numpy.org/}.
For the evaluation of 3D scene understanding, we use the Blender engine to generate 2D views of 3D scenes.
Existing datasets have also been used in creating~\bench, we use the images from~\citet{bitton2023vasr} for our evaluation on visual analogy, images from COCO~\citep{lin2014microsoft} for the counting evaluation, images from OpenFoodFact~\cite{openfoodfact} for the comparison evaluation, and ImageNet-R~\citep{hendrycks2020many} for the evaluation on attribute matching.
We have also sourced images from websites such as Pixsum, Pixabay, and arXiv for our evaluation.

\section{Qualitative Results}
\begin{figure}[t]
    \centering
    \includegraphics[width=.92\linewidth]{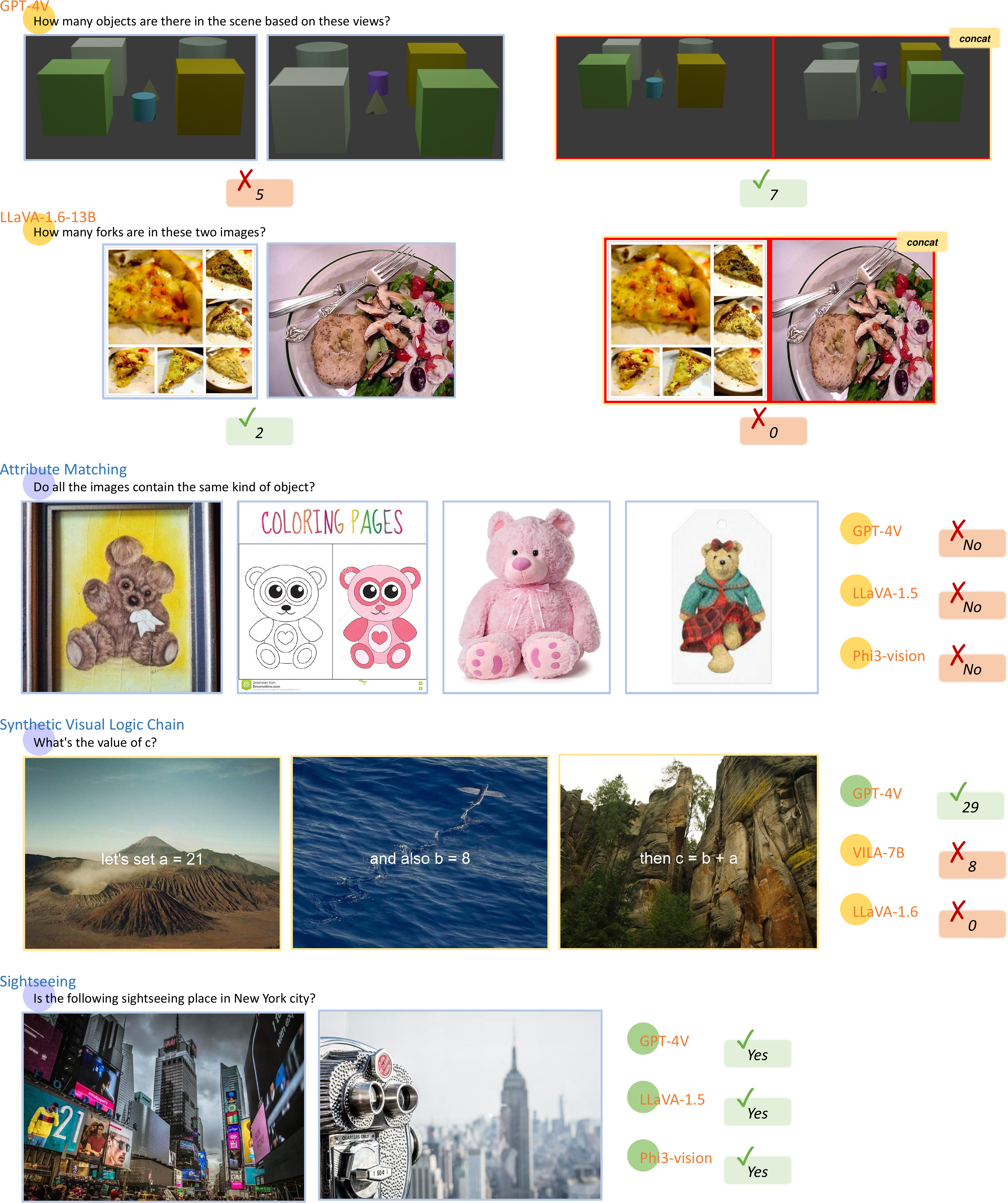}
    \caption{\textbf{Qualitative examples on~\bench.} No model can consistently solve our benchmark.}
    \label{fig:quality}
\end{figure}

When evaluated on~\bench, all current LVLMs show highly volatile performance. 
We showcase some qualitative examples in Figure~\ref{fig:quality}. 
Notably, the performance is vulnerable to how the images are presented, indicating that current models are unstable in solving multi-image tasks. 
We also find that the models struggle to collect useful information across images (e.g., the plain symbol content in synthetic logic chain questions).

\end{document}